\documentclass[preprint,review,12pt]{elsarticle}

\usepackage{color}
\usepackage{amssymb}
\usepackage{amsthm}
\usepackage{amsmath}
\usepackage{amssymb}
\usepackage{algorithm}
\usepackage{multirow} 
\usepackage{algorithmic}
\usepackage{subfig}
\newtheorem{theorem}{Theorem}
\renewcommand{\vec}[1]{\mathbf{#1}}
\journal{Neurocomputing}

\begin{document}

\begin{frontmatter}


\renewcommand{\vec}[1]{\mathbf{#1}}
\title{Clustering with Similarity Preserving}

 \author[label1]{Zhao Kang}\ead{zkang@uestc.edu.cn}
\author[label1]{ Honghui Xu}
\author[label2]{Boyu Wang}
\author[label3]{Hongyuan Zhu}
\author[label1]{ Zenglin Xu}
 \address[label1]{SMILE Lab, School of Computer Science and Engineering, University of Electronic Science and Technology of China, Chengdu 611731, China}
 \address[label2]{ GRASP Laboratory, Department of Computer and Information Science, University of Pennsylvania, PA 19104, USA}
 \address[label3]{Institute for Infocomm Research, A*Star, Singapore}

%

\begin{abstract}
Graph-based clustering has shown promising performance in many tasks. A key step of graph-based approach is the similarity graph construction. In general, learning graph in kernel space can enhance clustering accuracy due to the incorporation of nonlinearity. However, most existing kernel-based graph learning mechanisms is not similarity-preserving, hence leads to sub-optimal performance. To overcome this drawback, we propose a more discriminative graph learning method which can preserve the pairwise similarities between samples in an adaptive manner for the first time. Specifically, 
we require the learned graph be close to a kernel matrix, which serves as a measure of similarity in raw data.  Moreover, the structure is adaptively tuned so that the number of connected components of the graph is exactly equal to the number of clusters. Finally, our method unifies clustering and graph learning which can directly obtain cluster indicators from the graph itself without performing further clustering step. The effectiveness of this approach is examined on both single and multiple kernel learning scenarios in several datasets.
\end{abstract}


\end{frontmatter}


\section{Introduction}

Discovering clusters in unlabeled data is one of the most fundamental scientific tasks, with an endless list of practical applications in data mining, pattern recognition, and machine learning \cite{jain1999data,yang2017discrete,huang2018self,chen2012fgkm,ren2019semi}. It is well-known that labels are expensive to obtain, so clustering techniques are useful tools to process data and to reveal its underlying structure.

Over the past decades, a number of clustering techniques have been developed \cite{xu2005survey,huang2019auto,yang2018fast,peng2018integrate}. One main class of clustering methods is K-means and its various extensions. To some extent, these techniques are distance-based methods. K-means has been extensively investigated since its introduction in 1957 by Lloyd \cite{lloyd1982least}, due to its simplicity and effectiveness. However, it is only suitable for data points that are evenly spread around some centroids \cite{yang2017towards, chen2013twkm}. To make it work under general circumstances, much effort has been spent on mapping data to a certain space. One representative approach is using kernel method. The first Kernel K-means algorithm was proposed in 1998 \cite{scholkopf1998nonlinear}. Although some data points cannot be separated in the original data representation, they are linearly separable in kernel space.

Recently, robust Kernel K-means (RKKM) method has been developed \cite{du2015robust}. Different from other K-means algorithms, RKKM uses $\ell_{21}$-norm to evaluate the fidelity term. Consequently, RKKM can alleviate the adverse effects of noise and outliers considerably. It shows that RKKM can achieve superior performance on a number of real-world data sets. However, its performance still depends on the choice of the kernel function. 

Graph-based algorithms, as another main category of clustering methods, have been drawing growing attention. Among them, spectral clustering is a leading and highly popular method due to its ability in incorporating manifold information with good performance\cite{ng2002spectral,kang2018unified}. In particular, it embeds the data into the eigenspace of the Laplacian matrix, derived from the pairwise similarities between data points \cite{peluffo2016relationship}. A commonly used way of similarity measure is the Gaussian kernel \cite{von2007tutorial}. Nevertheless, it is challenging to select an appropriate scaling factor $\sigma$ \cite{zelnik2005self}. Kernel spectral clustering (KSC) \cite{alzate2010multiway} and its variants \cite{langone2016kernel} have also been proposed.

Recently, a novel approach which models graph construction as an optimization problem has been proposed \cite{nie2014clustering,Cheng2010,elhamifar2013sparse,liu2013robust,tang2018learning}. It works by either performing adaptive local structure learning or representing each data point as a weighted combination of other data points. The second approach can capture the global structure information and can be easily extended to kernel space \cite{kang2017kernel,kang2019low}. That is to say, one seeks to learn a high-quality graph from artificially constructed kernel matrix. These methods are free of similarity metrics or kernel parameters, thus they are more appealing to real-world applications. 

Although the above approach has shown much better performance than traditional methods, it also causes some information loss. In particular, it learns similarity graph from the data itself without considering other prior information. Consequently, some similarity information might get lost, which should be helpful for our graph learning \cite{haeffele2017structured,Kang2019aa}. On the other hand, preserving similarity information has been shown to be important for feature selection \cite{zhao2013similarity}. In \cite{zhao2013similarity}, new feature vector $f$ is obtained by maximizing $f^T\hat{K}f$, where $\hat{K}$ is the refined similarity matrix derived from original kernel matrix $K$. In this paper, we propose a way to preserve the similarity information between samples when we learn the graph and cluster labels. To the best of our knowledge, this is the first work that develops similarity preserving strategy for graph learning.

It is necessary to point out that the key point of this paper is the similarity preserving concept. Though there are many similarity learning methods in the literature, they often ignore to explicitly retain structure information of original data. Concretely, we expect our learned similarity matrix $Z$ approximates pre-defined kernel matrix $K$ to some extent. The quality of similarity matrix is crucial to many tasks, such as graph embedding \cite{cai2018comprehensive}, where the low-dimensional representation is expected to respect the neighborhood relation characterized by $Z$.

In addition, most existing graph-based clustering methods perform clustering in two separate steps \cite{ng2002spectral, liu2013robust, elhamifar2013sparse, kang2019robust}. Specifically, they first construct a graph. Then, the obtained graph is inputted to the spectral clustering algorithm. In this approach, the quality of the graph is not guaranteed, which might not be suitable for subsequent clustering \cite{nie2014clustering,kang2017twin}. In this paper, the structure information of the graph is explicitly considered in our model, so that the component number in the learned graph is equal to the number of clusters. Then, we can directly obtain cluster indicators from the graph itself without performing further graph cut or K-means clustering steps. Extensive experimental results validate the effectiveness of our proposed method.

The contributions of this paper are two-fold:
\begin{enumerate}
\item Our proposed model has the capability of similarity preserving. This is the first attempt to preserve the sample's similarity information when we construct the similarity graph. Consequently, the quality of the learned graph would be enhanced.
\item Cluster structure is seamlessly incorporated into our objective function. As a result, the component number in the learned graph is equal to the number of clusters, such that the vertices in each connected component of the graph are partitioned into one cluster. Therefore, we directly obtain cluster indicators from the graph itself without performing further graph cut or K-means clustering steps.
\end{enumerate}

\textbf{Notations.} Given a data matrix $X\in\mathcal{R}^{m \times n}$ with $m$ features and $n$ samples, we denote its $(i,j)$-th element and $i$-th column as $x_{ij}$ and $X_i$, respectively. The $\ell_2$-norm of vector \textbf{$x$} is represented by \textbf{$\|x\|=\sqrt{x^T\cdot x}$}, where $\textbf{$x^T$}$ is the transpose of \textbf{$x$}. The squared Frobenius norm is defined as $\|X\|_F^2=\sum_{ij}x_{ij}^2$. $I$ represents the identity matrix with the proper size. $ Z\geq 0$ means all the elements of $Z$ are nonnegative. $<\cdot,\cdot>$ denotes the inner product of two matrices.

\section{Preliminaries}
In this section, we give a brief overview of two popular similarity learning techniques which have been developed recently. 
\subsection{Adaptive Local Structure Learning}  
For each data point $X_i$, it can be connected to data point $X_j$ with probability $z_{ij}$. Closer points should have a larger probability, thus $z_{ij}$ characterizes the similarity between $X_i$ and $X_j$ \cite{niyogi2004locality,nie2014clustering}. Since $z_{ij}$ has the negative correlation with the distance between $X_i$ and $X_j$, the determination of $z_{ij}$ can be achieved by optimizing the following problem:
\begin{equation}
\begin{split}
&\min_{Z_i} \sum_{j=1}^n (\| X_i-X_j\|^2z_{ij}+\gamma z_{ij}^2), \\
&\quad\hspace{.1cm} s.t. \hspace{.2cm} Z_i^T\vec{1}=1, \hspace{.1cm}  0\leq z_{ij}\leq 1,
\end{split}
\label{local}
\end{equation}
where $\gamma$ is the trade-off parameter. Here, $Z$ is adaptively learned from the data. This idea has recently been applied in a number of problems. Nonnegative matrix factorization \cite{dacheng2017,huang2018adaptive}, feature selection \cite{du2015unsupervised}, multi-view learning \cite{nie2017multi}, just to name a few. One limitation of this method is that it can only capture the local structure information and thus the performance might be deteriorated. 

\subsection{Adaptive Global Structure Learning} 
To explore the global structure information, methods based on self-expression, have become increasingly popular in recent years \cite{Cheng2010,yang2014data}. The basic idea is to encode each datum as a weighted combination of other samples, i.e., its direct neighbors and reachable indirect neighbors. If $X_j$ is quite similar to $X_i$, coefficient $z_{ij}$, which denotes the contribution from $X_j$ to $X_i$, should be large. From this point of view, $z_{ij}$ can be viewed as the similarity between the data points. The corresponding optimization problem can be formulated as:
\begin{equation}
\min_{Z}\frac{1}{2} \|X-XZ\|_F^2+\gamma \|Z\|_F^2 \hspace{.2cm} s.t. \hspace{.2cm} Z\geq 0
\label{global}
\end{equation} 
This has drawn significant attention and achieved impressive performance in a number of applications, including face recognition \cite{zhang2011sparse}, motion segmentation \cite{liu2013robust,elhamifar2013sparse}, semi-supervised learning \cite{zhuang2017label}. 

As a matter of fact, Eq. (\ref{global}) is related to some dimension reduction methods. For example, in Locally Linear Embedding (LLE), $k-$nearest neighbors are first identified for each data point \cite{roweis2000nonlinear}. Then each data point is reconstructed by a linear combination of its k nearest neighbors. By contrast, Eq. (2) uses all data points and determines the neighbors automatically according to the optimization result. Thus, it is supposed to capture the global structure information. Eq. (2) is different from Locality Preserving Projections (LPP), which tries to preserve the neighborhood structure during the dimension reduction process \cite{he2004locality}. LPP uses a predefined similarity matrix to characterize the neighbor relations, while Eq. (2) is trying to learn this similarity matrix automatically from data. For Laplacian Eigenmaps (LE), a similarity graph matrix is also predefined \cite{belkin2002laplacian}. On the other hand, Principal Component Analysis (PCA) aims to find a projection so that the variance is maximized in low-dimensional space, which is less relevant to the similarity learning methods.

To capture the nonlinear structure information of data, Eq. (\ref{global}) can be easily extended to kernel space, which gives
\begin{equation}
\min_Z\hspace{.1cm} \frac{1}{2}Tr(K-2KZ+Z^TKZ)+\gamma \|Z\|_F^2 \hspace{.2cm} s.t. \hspace{.2cm} Z\geq 0,
\label{kernel}
\end{equation}
where $Tr(\cdot)$ is the trace operator and $K$ is the kernel matrix of $X$.
This model recovers the linear relations among the data in the new representation, and thus the nonlinear relations in the original space. Eq. (\ref{kernel}) is more general than Eq. (\ref{global}) and reduces to Eq. (\ref{global}) if a linear kernel function is applied.


\section{Similarity Preserving Clustering}
The aforementioned two learning mechanisms lead to much better performance than traditional similarity measure based techniques in many real-world applications. However, they ignore some important information. Specifically, as they operate on the data itself, some data relation information might get lost \cite{haeffele2017structured}. Since we seek to learn a high-quality similarity graph, data relation information would be crucial to our task. In this paper, we aim to retain this information.

Because the kernel matrix $K$ itself contains similarity information of data points, we expect $Z$ to be close to $K$. To this end, we optimize the following objective function
\begin{equation}
\max_Z <K,Z> \Leftrightarrow \max_Z Tr(KZ)\Leftrightarrow\min_Z -Tr(KZ).
\label{max}
\end{equation}
Although we claim similarity preserving, Eq. (\ref{max}) also keeps dissimilarity information. For example, if points $i$ and $j$ are from different clusters, $K_{ij}=K_{ji}=0$, then $Z_{ij}=Z_{ji}=0$ would hold. Note that we already have $-Tr(KZ)$ term in Eq. (\ref{kernel}). Hence we can combine Eq. (\ref{max}) and Eq. (\ref{kernel}) by introducing a coefficient $\alpha>1$, we have
\begin{equation}
\min_Z\hspace{.1cm} \frac{1}{2}Tr(K+Z^TKZ)-\alpha Tr(KZ)+\gamma \|Z\|_F^2 \hspace{.2cm} s.t. \hspace{.2cm} Z\geq 0.
\label{firststep}
\end{equation}
Although we just make a small modification to Eq. (\ref{kernel}), it makes a lot of sense in practice. By tuning parameter $\alpha$, we can control how much relation information we want to keep from the original kernel matrix. In particular, $\alpha$ can avoid the conflicts between the pre-computed similarity $K$ and the learned similarity $Z$. If $K$ is not suitable to reveal the underlying relationships among samples, we just set $\alpha=1$, which means that there is no similarity preserving effect. The influence of selecting parameter $\alpha$ is elaborated in Sec. \ref{parainfluence}.

Eq. (\ref{firststep}) provides a framework to learn graph matrix $Z$ with similarity preservation. Further clustering is achieved by using spectral clustering and K-means clustering on the learned graph. These separate steps often lead to suboptimal solutions \cite{nie2014clustering} and K-means is sensitive to the initialization of cluster centers. To this end, we propose to unify clustering with graph learning, so that two tasks can be simultaneously achieved. Speficially, if there are $c$ clusters in the data, we hope to learn a graph with exactly $c$ number of connected components. Obviously, Eq. (\ref{firststep}) can hardly satisfy such a constraint. To this end, we leverage the following theorem:
\begin{theorem}
\cite{mohar1991laplacian} The number of connected components $c$ is equal to the multiplicity of zero as an eigenvalue of its Laplacian matrix $L$.
\end{theorem}
Since $L$ is positive semi-definite, it has $n$ non-negative eigenvalues $\lambda_n\geq \cdots \lambda_2 \geq\lambda_1\geq0$. Theorem 1 indicates that if $\sum_{i=1}^c \lambda_i=0$ is satisfied, the graph $Z$ would be ideal and the data points are already clustered into $c$ clusters. According to Fan's theorem \cite{fan1949theorem}, we obtain
\begin{equation}
\sum_{i=1}^c \lambda_i=\min_{F\in\mathcal{R}^{n\times c}, F^TF=I} Tr(F^TLF),
\label{lap}
\end{equation}
where Laplacian matrix $L=D-Z$, $D$ is a diagonal matrix and its elements are the column sums of $Z$. Combining Eq.(\ref{lap}) with Eq.(\ref{firststep}), our proposed \textbf{S}imilarity \textbf{P}reserving \textbf{C}lustering (SPC) is formulated as
\begin{equation}
\begin{split}
\min_{Z,F}\hspace{.1cm}&\frac{1}{2} Tr(K\!+\!Z^TKZ)\!-\!\alpha Tr(KZ)\!+\!\beta Tr(F^TLF)+\gamma\|Z\|_F^2\\
&\quad s.t.\quad F^TF=I, \hspace{.1cm} Z\geq 0.
\end{split}
\label{obj}
\end{equation}
By solving Eq. (\ref{obj}), we can obtain a structured graph $Z$, which has exactly $c$ connected components. By running Matlab built-in function \emph{graphconncomp}, we can obtain which component each sample belongs to.

\subsection{Optimization}
The problem (\ref{obj}) can be easily solved with an alternating optimization approach. When $Z$ is fixed, Eq. (\ref{obj}) becomes
\begin{equation}
\min_F Tr(F^TLF)  \quad s.t. \quad F^TF=I.
\label{objf}
\end{equation}
It is quite standard to achieve $F$ which is formed by the $c$ eigenvectors of $L$ corresponding to the $c$ smallest eigenvalues.

When $F$ is fixed, Eq. (\ref{obj}) can be written column-wisely 
\begin{equation}
\min_{Z_i}\hspace{.1cm}\frac{1}{2}Z_i^TKZ_i-\alpha K_{i,:}Z_i+\frac{\beta}{2} d_i^T Z_i+\gamma Z_i^TZ_i,
\label{objz}
\end{equation}
where $d_i=\sum_{j=1}^n\|F_i-F_j\|^2$ and we have used equality $\sum_{i,j}\frac{1}{2}\|F_i-F_j\|^2z_{ij}=Tr(F^TLF)$. It is easy to achieve the closed-form solution
\begin{equation}
Z_i=(K+2\gamma I)^{-1}(\alpha K_{i,:}-\frac{\beta d_i}{2} )
\label{solz}
\end{equation}
Once parameter $\gamma$ is given, $(K+2\gamma I)$ becomes a constant. Therefore, we only perform the matrix inversion once. We summarize the steps in Algorithm 1. Our algorithm stops if the maximum iteration number 200 is reached or the relative change of $Z$ is less than $10^{-5}$.

\begin{algorithm}[!tb]
\caption{Similarity Preserving Clustering (SPC)}
\label{alg1}
 {\bfseries Input:} Kernel matrix $K$, parameters  $\alpha>1$, $\beta>0$, $\gamma>0$.\\
{\bfseries Initialize:} Random matrix $Z$.\\
 {\bfseries REPEAT}
\begin{algorithmic}[1]
 \STATE Update $F$ by solving (\ref{objf}).

   \STATE For each $i$, update the $i$-th column of $Z$ according to Eq. (\ref{solz}).
\STATE Project $Z$ by $Z=\textit{max}(Z,0)$.
\end{algorithmic}
\textbf{ UNTIL} {stopping criterion is met.}
\end{algorithm}

\section{Multiple Kernel Learning}
 Different kernels correspond to different notions of similarity and lead to different results. This makes it not be reliable for practical applications. Multiple Kernel Learning (MKL) offers a principal way to encode complementary information and automatically learning an optimal combination of distinct kernels \cite{sonnenburg2006large,kang2018self}. Instead of heuristic kernel selection, a principled method is developed to automatically learn a good combination of multiple kernels.

 Specifically, suppose there are in total $r$ kernels, we introduce kernel weight for each kernel, e.g., $w_i$ for kernel $K^i$. We denote the combined kernel as $H=\sum_{i=1}^r w_iK^i$, and the weight distribution satisfies $\sum_{i=1}^r\sqrt{w_i}=1$ \cite{gonen2011multiple}. Finally, our multiple kernel learning based similarity preserving clustering (mSPC) method can be formulated as 
\begin{equation}
\begin{split}
\min_{Z,F,w}\hspace{.1cm}&\frac{1}{2} Tr(H\!+\!Z^THZ)\!-\!\alpha Tr(HZ)\!+\!\beta Tr(F^TLF)\\
&+\gamma\|Z\|_F^2\quad
s.t.\quad F^TF=I,\quad Z\geq 0.\\
&\hspace{.2cm}H=\sum\limits_{i=1}^r w_iK^i,\hspace{.2cm} \sum\limits_{i=1}^r \sqrt{w_i}=1,\hspace{.2cm} w_i\ge 0.
\end{split}
\label{multimodel}
\end{equation}
The problem (\ref{multimodel}) can be solved in a similar way as (\ref{obj}). In specific, we repeat the following steps.

1) Updating $Z$ and $F$ when $\vec{w}$ is fixed: We can directly obtain $H$, and the optimization problem (\ref{multimodel}) is identical to Eq. (\ref{obj}). We implement Algorithm 1 with $H$ as the input kernel matrix.

2) Updating $\vec{w}$ when $Z$ and $F$ are fixed: Solving Eq. (\ref{multimodel}) with respect to $\vec{w}$ can be reformulated as
\begin{equation}
\label{optie}
\min_\vec{w} \sum\limits_{i=1}^r w_i h_i   \quad s.t.\quad  \sum\limits_{i=1}^r \sqrt{w_i}=1, \hspace{.1cm} w_i\ge 0, 
\end{equation}
where 
\begin{equation}
\label{h}
h_i=Tr(K^i-2\alpha K^iZ+Z^TK^iZ).
\end{equation}
The Lagrange function of Eq. (\ref{optie}) is 
\begin{equation}
\mathcal{J}(\vec{w})=\vec{w}^T\vec{h}+\eta (1-\sum_{i=1}^r\sqrt{w_i}).
\end{equation}
By utilizing the Karush-Kuhn-Tucker (KKT) condition with $\frac{\partial \mathcal{J}(\vec{w})}{\partial w_i}=0$ and the constraint $\sum\limits_{i=1}^r \sqrt{w_i}=1$, we have $\vec{w}$ as follows:
\begin{equation}
\label{weight}
w_i=(h_i \sum_{j=1}^r \frac{1}{h_j})^{-2}.
\end{equation}

\begin{algorithm}
\caption{The algorithm of mSPC}
\label{alg2}
 {\bfseries Input:} A set of kernel matrices $\{K^i\}_{i=1}^r$, parameters  $\alpha>1$, $\beta>0$, $\gamma>0$.\\
{\bfseries Initialize:} Random matrix $Z$, $w_i=1/r$.\\
 {\bfseries REPEAT}
\begin{algorithmic}[1]
\STATE Calculate $H$.
 \STATE Update $F$ by performing singular value decomposition on $L=D-Z$ and finding the $c$ smallest eigenvectors.
\STATE Update $Z$ column-wisely according to (\ref{solz}).
\STATE Update $\vec{h}$ by (\ref{h}).
\STATE Calculate $\vec{w}$ according to (\ref{weight}).
\end{algorithmic}
\textbf{ UNTIL} {stopping criterion is met.}
\end{algorithm}

\section{Experiment}
In this section, we perform extensive experiments to demonstrate the effectiveness of our proposed models.

\begin{figure}[!htbp]
\centering
\subfloat[k-means]{\includegraphics[width=7cm,height=4cm]{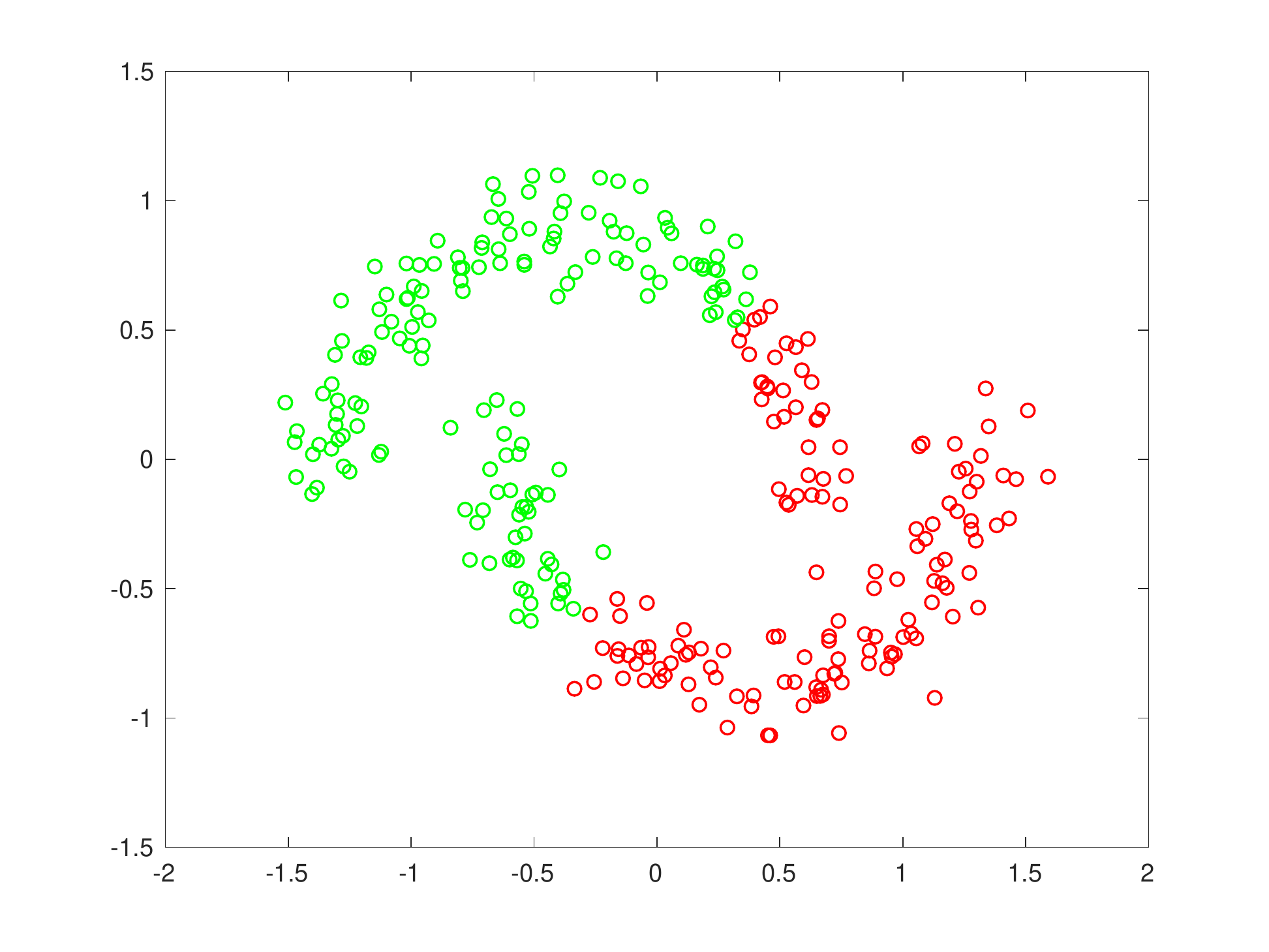}}
\subfloat[SPC]{\includegraphics[width=7cm,height=4cm]{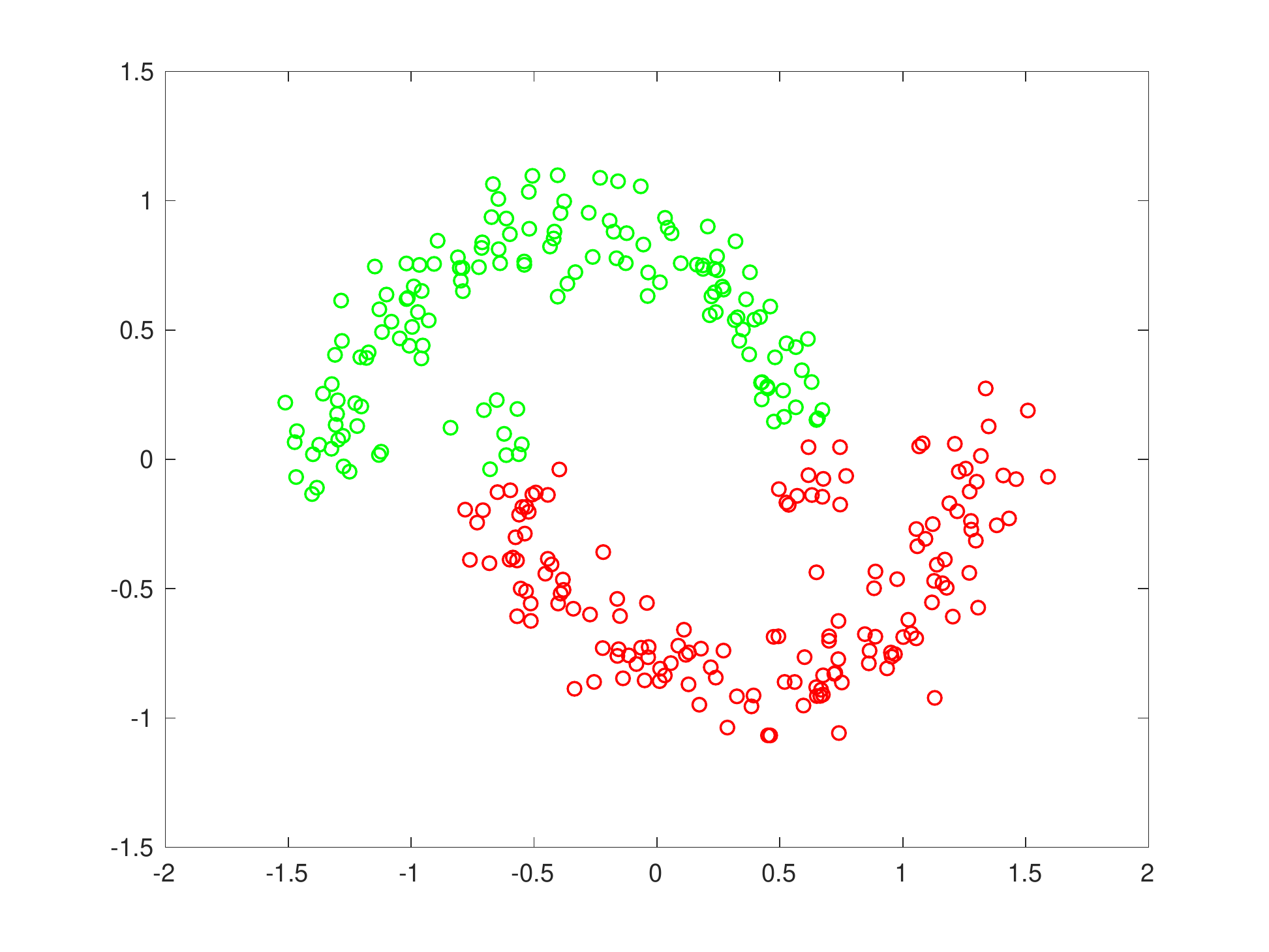}}
\caption{The clustering results on synthetic data.}
\label{syn}
\end{figure}
\subsection{Experiment on Synthetic Data}
We generate a synthetic data set with 300 points. The data points distribute in the pattern of two moons. Each moon is considered as a cluster. In Figure \ref{syn}, we present the clustering results of our proposed SPC and standard k-means. Gaussian kernel with $t=10$ is used in our SPC model. We can observe that our method performs much better than k-means. We quantitatively assess the clustering performance in terms of accuracy (Acc), normalized mutual information (NMI), and Purity. For SPC, Acc, NMI, and Purity are 93\%, 63.49\%, 93\%, respectively. Correspondingly, k-means produces 73.67\%, 16.87\%, and 73.67\%.
\subsection{Experiment on Real Data}
\subsubsection{Data sets}
\begin{table}[!htbp]
\centering
\caption{Description of the data sets}
\label{data}
\renewcommand{\arraystretch}{.8}
\begin{tabular}{|l|c|c|c|}
\hline
&\textrm{\# instances}&\textrm{\# features}&\textrm{\# classes}\\\hline
\textrm{YALE}&165&1024&15\\\hline
\textrm{JAFFE}&213&676&10\\\hline
\textrm{ORL}&400&1024&40\\\hline
\textrm{YEAST}&1484&1470&10\\\hline
\textrm{USPS}&1854&256&20\\\hline
\textrm{TR11}&414&6429&9\\\hline
\textrm{TR41}&878&7454&10\\\hline
\textrm{TR45}&690&8261&10\\\hline
\end{tabular}
\end{table}
\begin{figure*}[!htbp]
\centering
\subfloat[JAFFE\label{jaffe}]{\includegraphics[width=.33\textwidth]{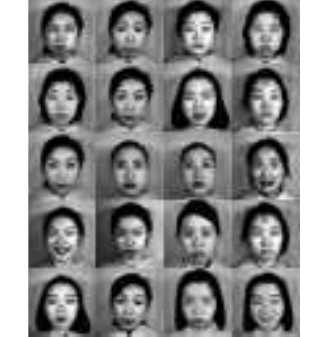}}
\subfloat[YALE\label{yale}]{\includegraphics[width=.33\textwidth]{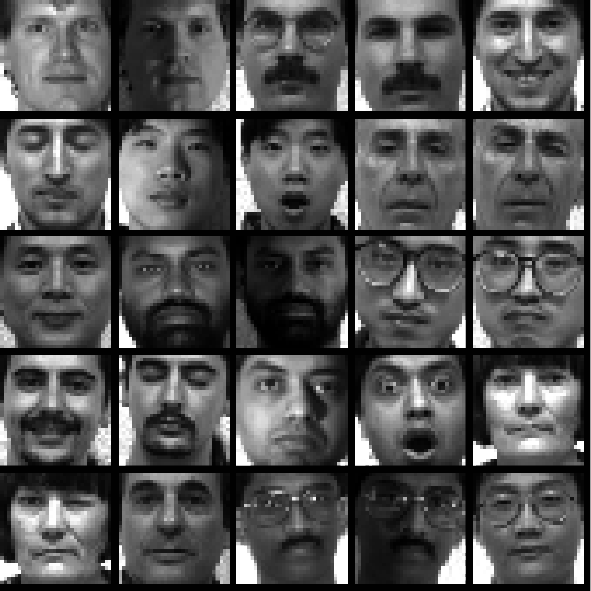}}
\subfloat[USPS\label{usps}]{\includegraphics[width=.33\textwidth]{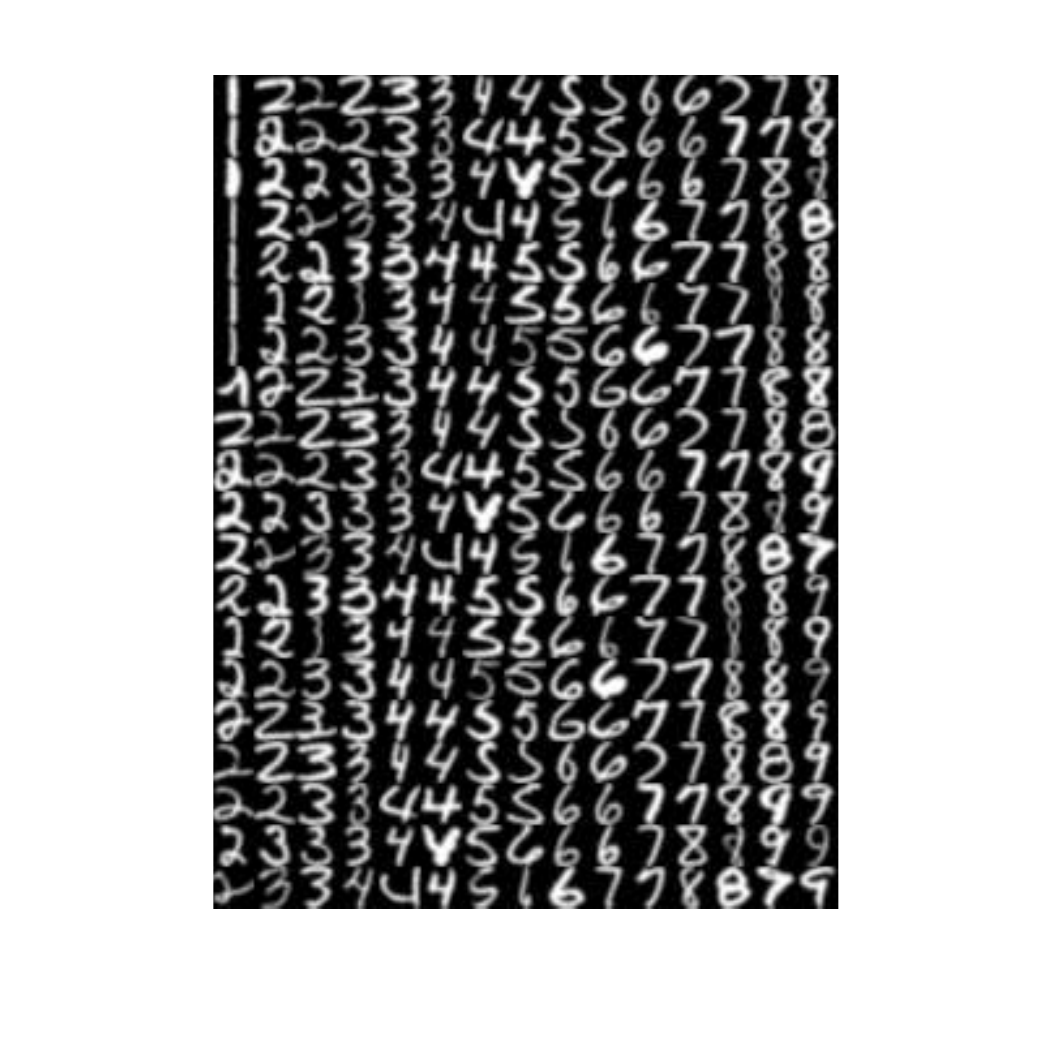}}
\caption{Sample images of USPS, JAFFE, and YALE.}
\end{figure*}

We conduct our experiments with eight benchmark data sets, which are widely used in clustering experiments. We show the statistics of these data sets in Table \ref{data}.

These data sets are from different fields. Specifically, YALE\footnote{http://vision.ucsd.edu/content/yale-face-database}, JAFFE\footnote{http://www.kasrl.org/jaffe.html}, ORL\footnote{http://www.cl.cam.ac.uk/research/dtg/attarchive/facedatabase.html} are three face databases. Each image represents different facial expressions or configurations due to times, illumination conditions, and glasses/no glasses. Figures \ref{jaffe} and \ref{yale} show some example images from JAFFE and YALE database, respectively. YEAST is a microarray data set. USPS data set\footnote{http://www-stat.stanford.edu/~tibs/ElemStatLearn/data.html} is obtained from the scanning of handwritten digits from envelopes by the U.S. Postal Service. Some sample digits are shown in Figure \ref{usps}. The last three data sets in Table \ref{data} are text data\footnote{http://www-users.cs.umn.edu/~han/data/tmdata.tar.gz}.

We manually construct 12 kernels. They consist of seven Gaussian kernels  $K(x,y)=exp(-\|x-y\|_2^2/(td_{max}^2))$ with $t\in\{0.01, 0.05, 0.1, 1, 10, 50, 100\}$ and $d_{max}$ denotes the maximal distance between data points; four polynomial kernels $K(x,y)=(a+x^Ty)^b$ of the form with $a\in\{0,1\}$ and $b\in\{2,4\}$; a linear kernel $K(x,y)=x^Ty$. Furthermore, all kernel matrices are normalized to $[0,1]$ range to avoid numerical inconsistence.

\begin{table*}[!htbp]
\centering
\renewcommand{\arraystretch}{1.}
\setlength{\tabcolsep}{.2pt}
\subfloat[Accuracy(\%)\label{acc}]{
\resizebox{1.\textwidth}{!}{

\begin{tabular}{ |l  |c| c| c|c |c|  c|c| |c| c| c |c| c| c| c| c| c|c }
\hline
Data 		   & SC		      & RKKM	 &SSR &CAN    & SPC1	&SPC	&KSC	        & MKKM  & AASC  & RMKKM & mSPC1 		      &mSPC		       	   \\	\hline
YALE  	 & 49.42(40.52) & 48.09(39.71)&54.55&58.79 & 55.29(45.07)  &\textbf{60.60}(46.95) &36.36(29.76)    & 45.70 & 40.64 & 52.18 & 56.97            &\textbf{63.03}  \\	\hline
JAFFE 	 & 74.88(54.03) & 75.61(67.98)&87.32&98.12  &97.18(86.55)&\textbf{98.03}(86.20)&72.77(66.48)  & 74.55 & 30.35 & 87.07 &97.18  &\textbf{98.12}             \\	\hline
ORL    & 57.96(46.65) & 54.96(46.88)&69.00&61.50  & 62.47(50.64)    &\textbf{75.75}(52.48)&37.00(32.70)   & 47.51 & 27.20 & 55.60 & 65.25            & \textbf{75.43}  	\\	\hline
YEAST&35.55(30.89)&34.04(31.52)&29.99&34.62&35.72(30.55)&\textbf{37.85}(31.65)&31.19(28.36)&13.04&35.38&31.63&36.08&\textbf{39.15}	\\	\hline
USPS&35.18(26.90)&65.16(55.72)&64.83&62.84&65.94(56.38)&\textbf{67.25}(56.94)&61.49(47.87)&63.72&37.36&65.47&65.32&\textbf{68.74}	\\	\hline
TR11    & 50.98(43.32) & 53.03(45.04) &41.06&38.89 & 70.88(54.25) & \textbf{78.26}(56.37)&50.73(41.86)  & 50.13 & 47.15 & 57.71 & 73.43   &\textbf{79.63}             	\\	\hline
TR41  & 63.52(44.80) & 56.76(46.80) &63.78&62.87 & 67.28(52.75)&\textbf{72.89}(49.40) &53.42(45.11)   & 56.10 & 45.90 & 62.65 & 67.31	  & \textbf{80.41}           	\\	\hline
TR45 & 57.39(45.96) & 58.13(45.69) &71.45&56.96 & 73.59(53.06)  & \textbf{75.07}(57.26)&53.48(44.99)        & 58.46 & 52.64 & 64.00 & 74.35      	  & \textbf{75.64} \\	\hline
\end{tabular}
}

}\\
\renewcommand{\arraystretch}{1.}
\subfloat[NMI(\%)\label{NMI}]{
\resizebox{1.\textwidth}{!}{
\begin{tabular}{ |l  |c| c| c|c |c|  c| c||c| c| c |c| c|c|  c| c| c|c  }
\hline
Data 	    & SC	       	& RKKM	  &SSR   &CAN    & SPC1		&SPC	&KSC        & MKKM  & AASC  & RMKKM & mSPC1			&mSPC		          \\	\hline
YALE 	  & 52.92(44.79)  & 52.29(42.87)&57.26&57.67  & 56.37(45.00)& \textbf{61.32}(45.62)  &44.84(35.47)      & 50.06 & 46.83 & 55.58 & 56.52         & \textbf{61.36}     	\\	\hline
JAFFE 	  & 82.08(59.35)  & 83.47(74.01)&92.93&97.31  &96.35(84.67)&\textbf{98.62}(83.30)&73.67(68.39)    & 79.79 & 27.22 & 89.37 & 95.61& \textbf{ 97.36}               	\\	\hline
ORL 	& 75.16(66.74)  & 74.23(63.91) &84.23&76.59 & 79.36(63.98) & \textbf{86.06}(66.56) & 56.78(54.21)      & 68.86 & 43.77 & 74.83 & 80.04	        & \textbf{85.93}  	\\	\hline
YEAST &\textbf{21.38}(6.18)&17.27(9.31)&15.85&20.06&15.37(9.62)&16.44(10.25)&13.87(13.83)&10.29&\textbf{21.19}&20.71&15.89&16.23	\\	\hline
USPS &29.71(21.33)&63.94(52.90)&72.68&76.13&74.85(55.28)&\textbf{78.54}(58.93)&48.45(40.27)&62.25&29.81&63.60&75.29&\textbf{79.88}	\\	\hline
TR11    & 43.11(31.39)  & 49.69(33.48) &27.60&19.17 & 58.14(37.42)  & \textbf{63.10}(35.94)&46.22(36.51)       & 44.56 & 39.39 & 56.08 & 60.15       	&  \textbf{63.90}  	\\	\hline
TR41   & 61.33(36.60)  & 60.77(40.86)&59.56 &51.13 & 65.90(43.28)&    \textbf{71.22}(38.54)   & 44.21(39.19)      & 57.75 & 43.05 & 63.47 & 65.11        	& \textbf{70.50}   	\\	\hline
TR45  & 48.03(33.22)  & 57.86(38.96)&67.82 &49.31 & 74.21(44.29)   & \textbf{75.94}(46.28) & 43.41(42.95)    & 56.17 & 41.94 & 62.73 & \textbf{74.97}& 74.57             	\\	\hline
\end{tabular}
}
}\\
\renewcommand{\arraystretch}{1.}
\subfloat[ Purity(\%)\label{purity}]{
\resizebox{1.\textwidth}{!}{
\begin{tabular}{ |l  |c| c| c|c |c|  c| c||c| c| c |c| c| c| c| c| c|c }\hline
Data  	   & SC	 	      & RKKM	&SSR&CAN	  & SPC1	&SPC	&KSC	        & MKKM  & AASC  & RMKKM & mSPC1 		        &mSPC		     		\\	\hline
YALE   	 & 51.61(43.06) & 49.79(41.74)&58.18 &59.39 & 56.79(55.25)  & \textbf{60.53}(56.28) &44.85(34.67)        & 47.52 & 42.33 & 53.64 & 60.00	            & \textbf{66.67}	\\	\hline
JAFFE  	 & 76.83(56.56) & 79.58(71.82)&96.24 &98.12 & 97.85(96.03) &\textbf{98.25}(97.02) &77.00(72.30)    & 76.83 & 33.08 & 88.90 & 97.18	& \textbf{98.12}        	\\	\hline
ORL   	& 61.45(51.20) & 59.60(51.46) &76.50&68.5 & 73.28(70.02) & \textbf{82.08}(76.56) &41.00(36.73)        & 52.85 & 31.56 & 60.23 & 77.00	&    \textbf{82.69}      	\\	\hline
YEAST &53.05(35.37)&45.39(38.18)&44.29&58.97&57.38(40.83)&\textbf{65.72}(44.75)&32.21(31.14)&32.58&52.71&33.21&60.27&\textbf{66.51}	\\	\hline
USPS &37.48(33.27)&72.49(62.49)&75.84&71.89&76.90(63.78)&\textbf{77.54}(64.82)&62.84(51.48)&70.77&42.40&73.45&77.03&\textbf{79.25}	\\	\hline
TR11   & 58.79(50.23) & 67.93(56.40)&85.02&44.20  & 81.79(80.12) & \textbf{90.10}(83.86)  & 52.90(46.76)       & 65.48 & 54.67 & 72.93 & 87.44	& \textbf{93.04}      	\\	\hline
TR41  	 & 73.68(56.45) & 74.99(60.21)&75.40&67.54& 73.05(71.13)& \textbf{80.67}(74.79)&53.42(47.92)     & 72.83 & 62.05 & 77.57 & 73.69	& \textbf{82.45}			\\	\hline
TR45   & 61.25(50.02) & 68.18(53.75) &83.62&60.87 & 78.74(77.82)& \textbf{86.32}(80.03) &  55.51(49.29)      & 69.14 & 57.49 & 75.20 & 78.26	&\textbf{87.59}        	\\	\hline
\end{tabular}
}
}
\caption{Clustering results measured on benchmark data sets. The average performance on those 12 kernels are put in parenthesis. For KSC, we run 10 times and report the best performance and their mean value. The best results for single and multiple kernel methods are highlighted in boldface. \label{clusterres}}
\end{table*}

\subsubsection{Comparison Methods}
To fully investigate the performance of our method on clustering, we choose a good set of methods to compare. In general, they can be classified into two categories: graph-based and kernel-based clustering methods.
\begin{itemize}
\item{\textbf{Spectral Clustering (SC) }\cite{ng2002spectral}. We use kernel matrix as its graph  input. For our SPC, we learn graph from kernels.}
\item{\textbf{Robust Kernel K-means (RKKM)}\footnote{https://github.com/csliangdu/RMKKM}\cite{du2015robust}. As an extension to the classical K-means clustering method, RKKM has the capability of dealing with nonlinear structures, noise, and outliers in the data. We also compare with its multiple kernel learning version: RMKKM. }
\item{\textbf{Simplex Sparse Representation (SSR)} \cite{huang2015new}. Based on self-expression, SSR achieves satisfying performance in numerous data sets. }
\item{\textbf{Clustering with Adaptive Neighbor (CAN)} \cite{nie2014clustering}. Based on adaptive local structure learning, CAN constructs the similarity graph by Eq. (\ref{local}).  }
\item{\textbf{Kernel Spectral Clustering (KSC)} \cite{alzate2010multiway}. Based on a weighted kernel principal component analysis strategy, KSC performs multiway spectral clustering. Moreover, Balanced Line Fit (BLF) is proposed to obtain model parameters.} 
\item{Our proposed \textbf{SPC} and \textbf{mSPC}\footnote{https://github.com/sckangz/SPC}. Our proposed single kernel and multiple kernel learning based similarity preserving clustering methods.}
\item{\textbf{SPC1} and \textbf{mSPC1}. To observe the effect of similarity preserving, we let $\alpha=1$ in SPC and name this method as SPC1. Similarly, we have mSPC1. They are equivalent to the methods in \cite{kang2017twin}.}  
\item{\textbf{Multiple Kernel K-means (MKKM)}\footnote{http://imp.iis.sinica.edu.tw/IVCLab/research/Sean/mkfc/code}\cite{huang2012multiple}. The MKKM  extends K-means in a multiple kernel setting. It imposes a different constraint on the kernel weight distribution. }

\item{\textbf{Affinity Aggregation for Spectral Clustering (AASC)}\footnote{http://imp.iis.sinica.edu.tw/IVCLab/research/Sean/aasc/code}\cite{huang2012affinity}. The AASC is an extension of spectral clustering to the situation when multiple affinities exist. }
\end{itemize}

For a fair comparison, we either use the recommended parameter settings in their respective papers or tune each method to obtain the best performance. In fact, the optimal performance for SC, RKKM, MKKM, AASC, and RMKKM methods can be easily obtained through implementing the package given in \cite{du2015robust}. SC, SSR, and CAN are parameter-free models. KSC selects parameters based on Balanced Line Fit principle. 

\begin{figure}[!htbp]
\centering
\subfloat[\label{diffdz}]{\includegraphics[width=7cm,height=4cm]{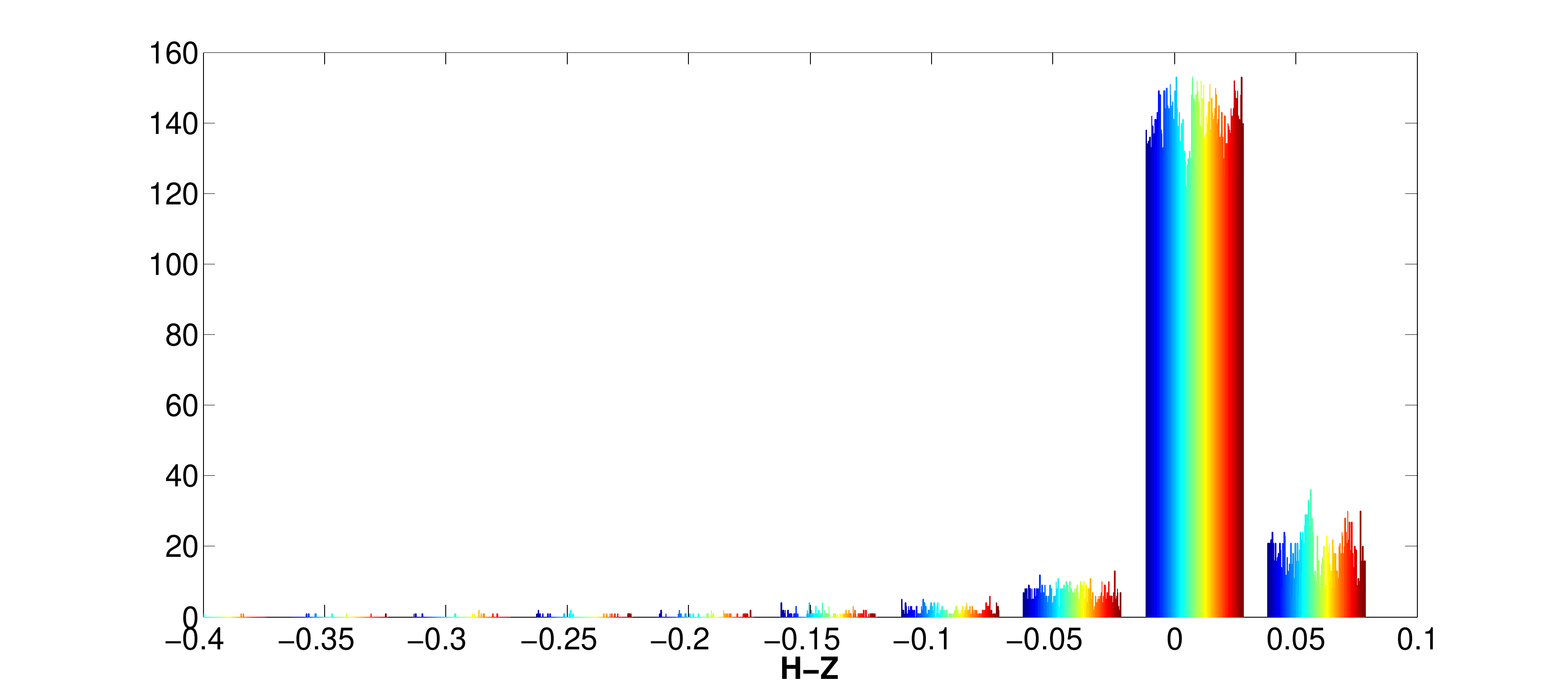}}
\subfloat[\label{diffxz}]{\includegraphics[width=7cm,height=4cm]{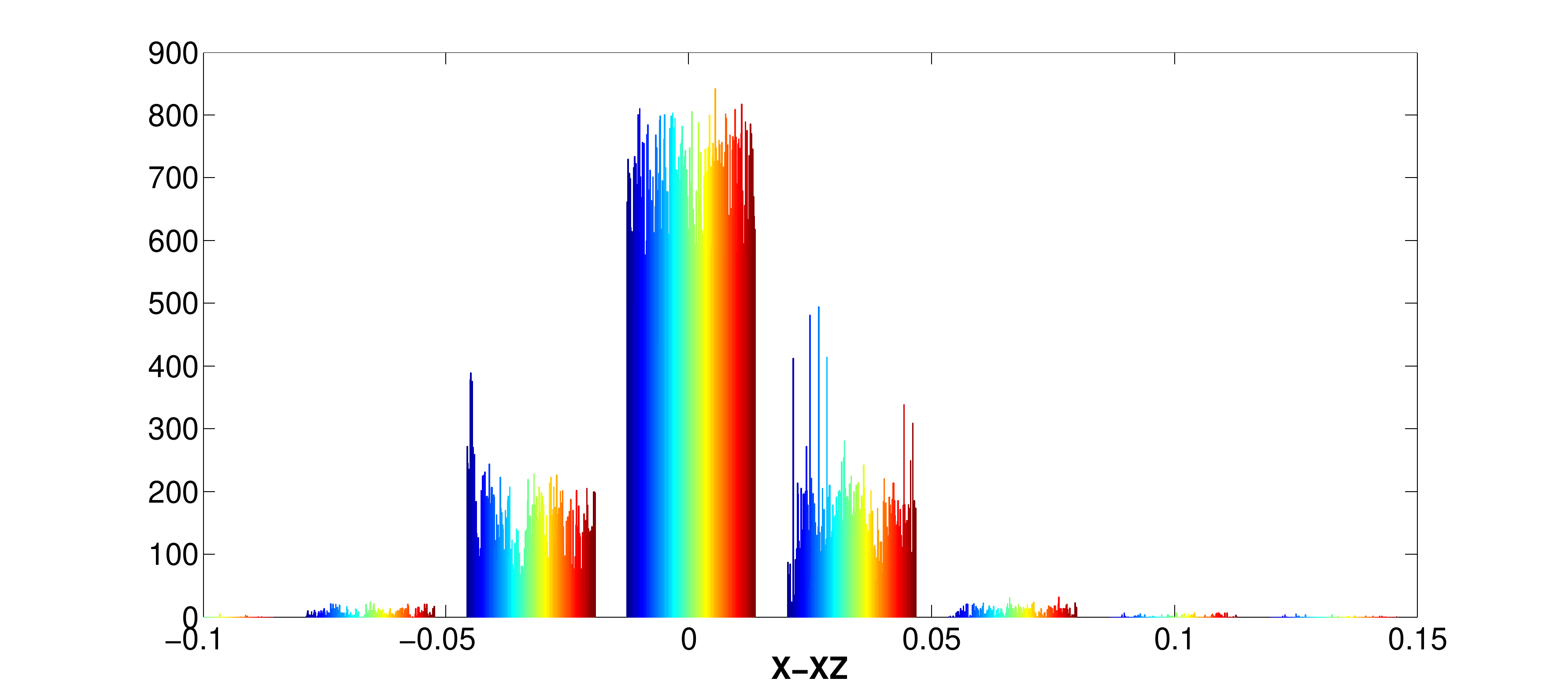}}
\caption{The visualization of similarity preserving effect.}
\label{diff}
\end{figure}

\subsubsection{Results}
All results are summarized in Table \ref{clusterres}. We can see that our methods SPC and mSPC outperform others in most cases. In particular, we have the following observations: 1) The improvement of SPC over SPC1 is considerable. Noted that the only difference between SPC and SPC1 is that SPC explicitly considers the similarity preserving effect. In other words, SPC adds the proposed term Eq. (\ref{max}), which aims to keep the learned graph matrix $Z$ close to the kernel matrix $K$, so that the similarity information carried by the kernel matrix will transfer to the learned graph matrix. Hence this demonstrates the significance of similarity preserving in graph learning; 2) For multiple kernel methods, mSPC also performs better than mSPC1    
in most experiments. This once again confirms the importance of similarity preserving; 3) Compared to self-expression based method SSR, our advantage is also obvious. For example, in TR11, SPC enhances the accuracy from 41.06\% to 78.26\%. Note that our basic objective function Eq. (\ref{kernel}) is also derived from self-expression idea. However, our method is kernel method; 4) With respect to traditional spectral clustering, kernel spectral clustering, the recently proposed robust kernel K-means method, adaptive local structure graph learning method, the improvement is very promising; 5) In terms of multiple kernel learning approach, mSPC also achieves much better performance than other state-of-the-art techniques.  

To better illustrate the effect of similarity preserving, we visualize the results of YALE data in Figure \ref{diff}. In specific, Figure \ref{diffdz} plots the histogram of $H-Z$, i.e., the difference between the learned kernel in Eq. (\ref{multimodel}) and similarity matrix. We can see that they are quite close for most elements and the difference is the refinement brought by our learning algorithm. The manually constructed kernel matrix often fails to reflect the underlying relationships among samples due to the inherent noise or the inappropriate use of a metric function. This is validated by the experimental results. Note that for SC method, we directly treat kernel matrix as similarity matrix, while for our proposed SPC method, we use the learned similarity matrix $Z$ to perform clustering. It can be seen that the results of SPC are much better than that of SC. 

Figure \ref{diffxz} displays the difference between the original data $X$ and the reconstructed data $XZ$. Good reconstruction means that $Z$ represents the similarity pretty well. The reconstruction error accounts for noise or outliers in the original data. As shown by Figure \ref{diffxz}, our learned $Z$ reconstructs the original data with a small error. Therefore, our proposed approach can achieve a high-quality similarity matrix.

\subsubsection{Parameter Analysis}
\label{parainfluence}
\begin{figure*}[!htbp]
\centering
\subfloat[$\alpha=10$\label{yale10b}]{\includegraphics[width=.33\textwidth]{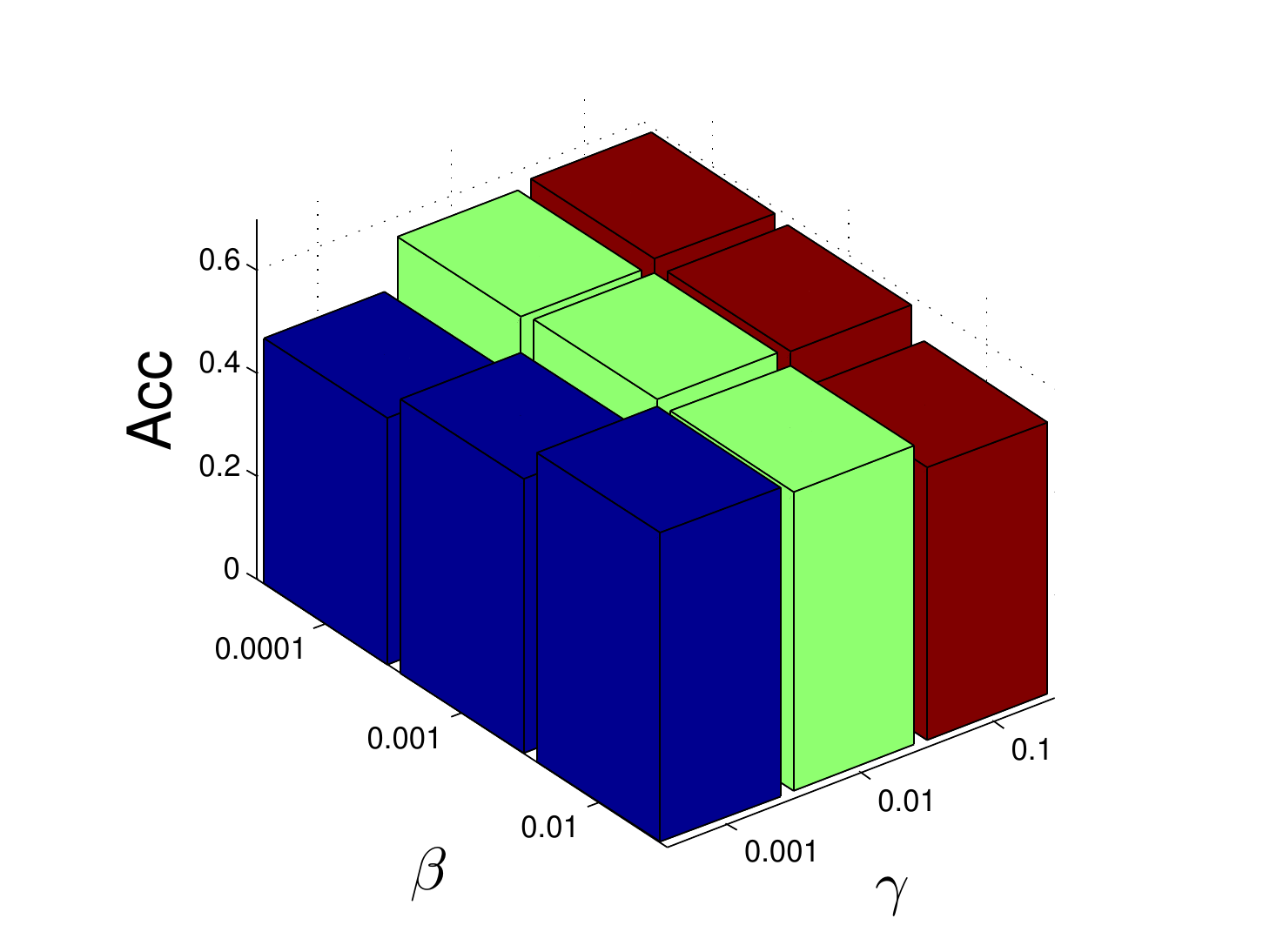}}
\subfloat[$\alpha=10$]{\includegraphics[width=.33\textwidth]{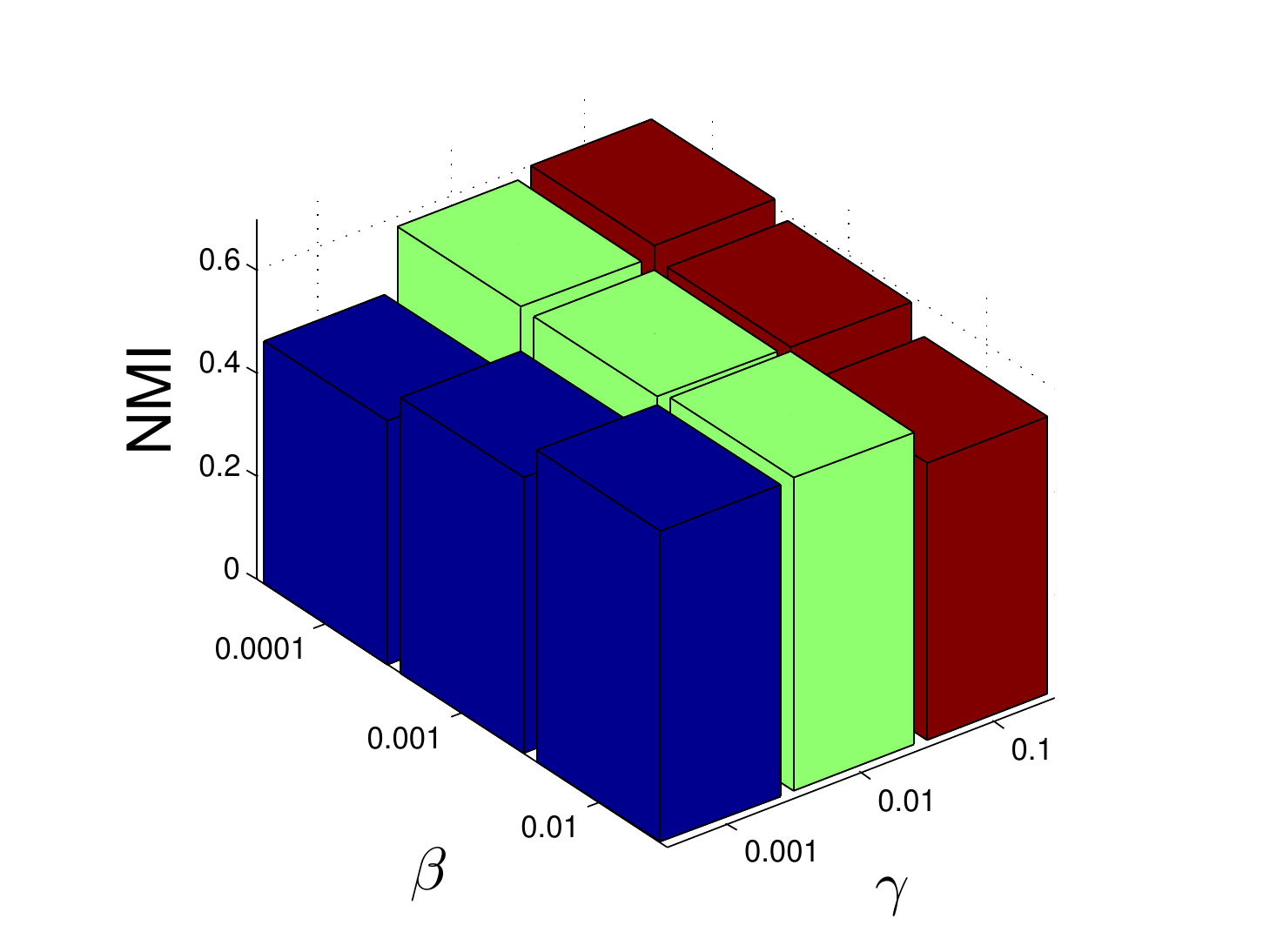}}
\subfloat[$\alpha=10$\label{yale10}]{\includegraphics[width=.33\textwidth]{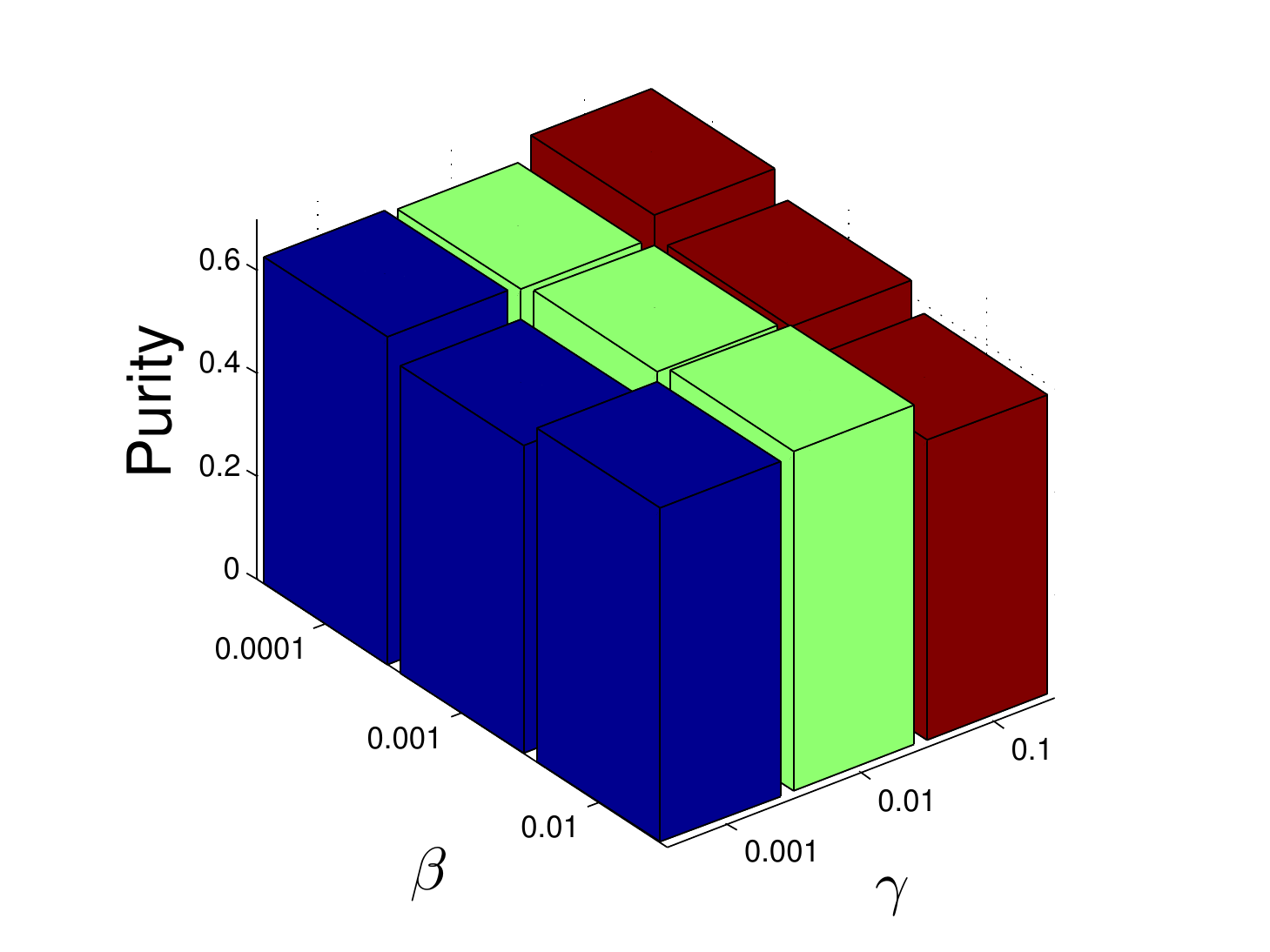}}\\
\subfloat[$\alpha=100$\label{yale100b}]{\includegraphics[width=.33\textwidth]{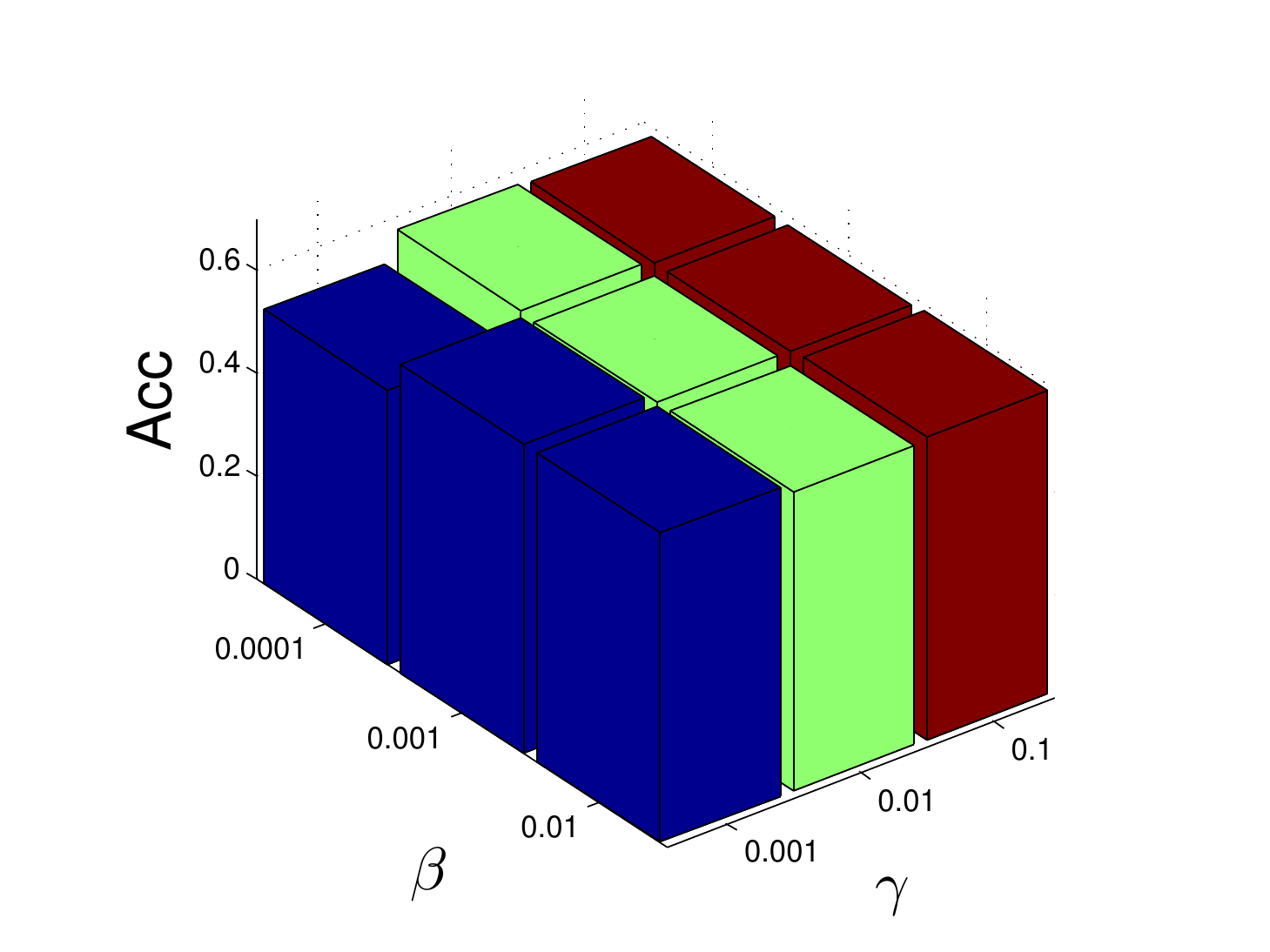}}
\subfloat[$\alpha=100$]{\includegraphics[width=.33\textwidth]{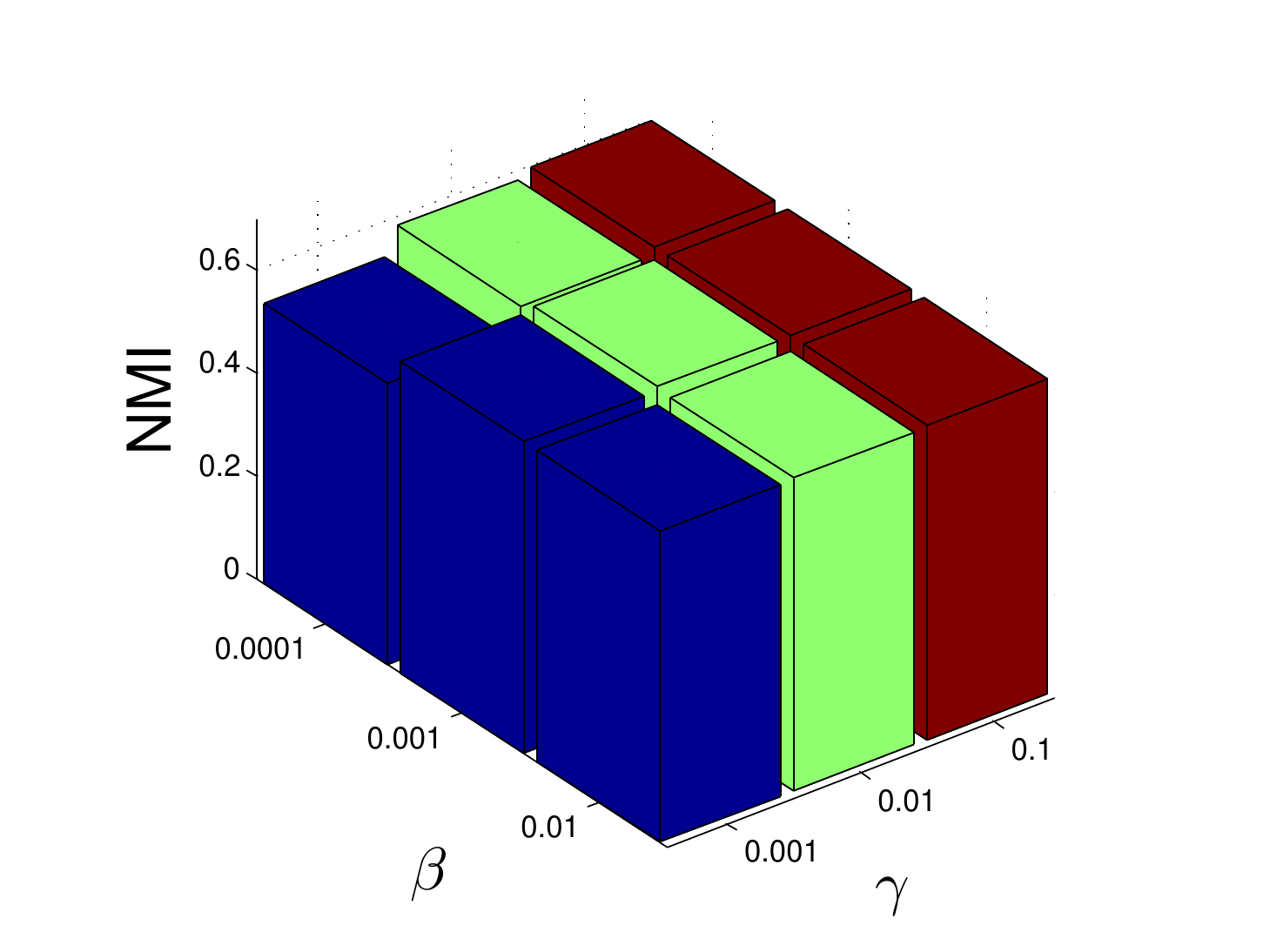}}
\subfloat[$\alpha=100$\label{yale100}]{\includegraphics[width=.33\textwidth]{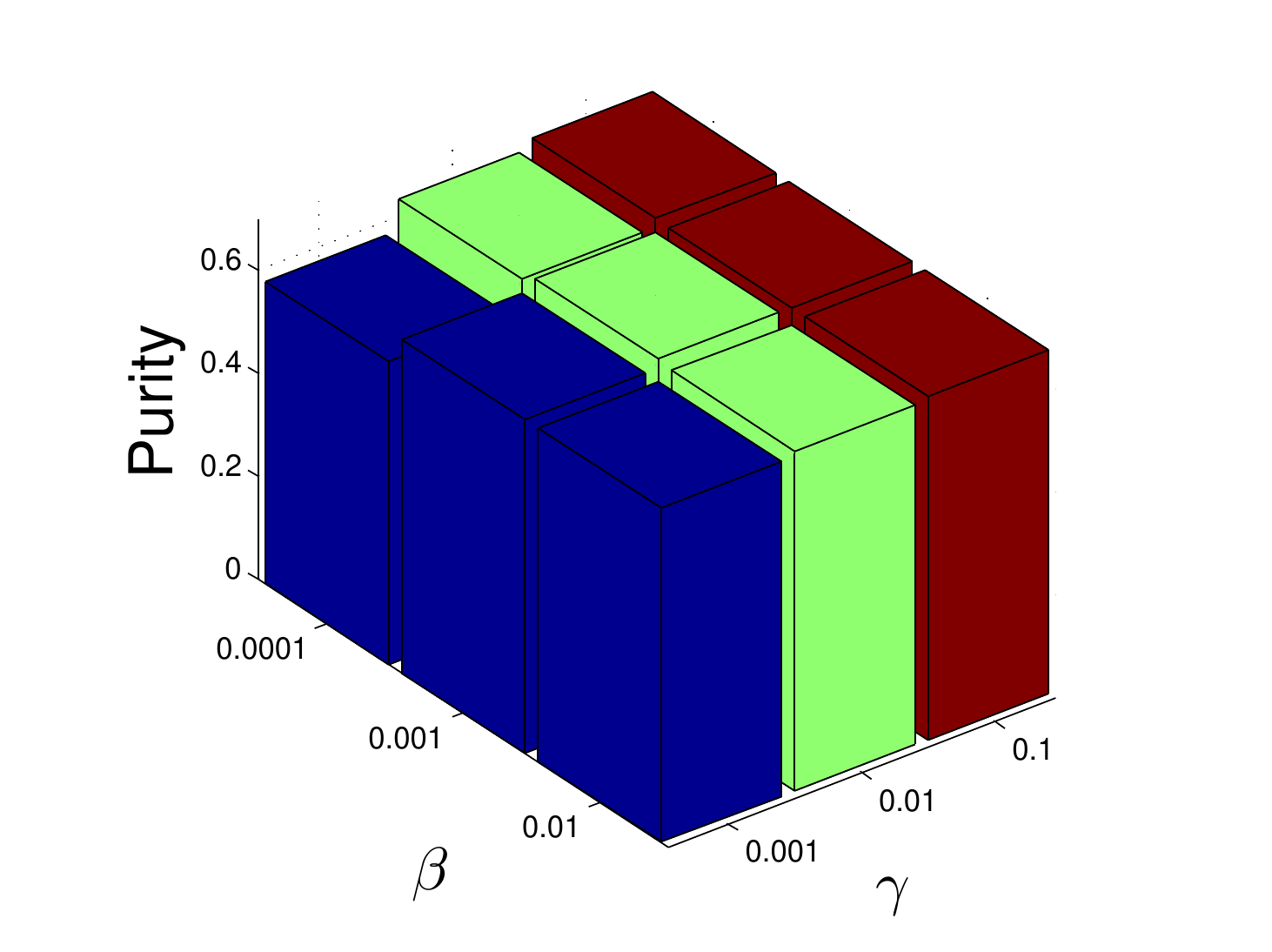}}
\caption{The influence of parameters on YALE data set.}
\label{yaleacc}
\end{figure*}

As shown in Eq. (\ref{multimodel}), there are three parameters in our model. As we mentioned previously, $\alpha$ is bigger than one. Take YALE data set as an example, we demonstrate the sensitivity of our model mSPC in Figure \ref{yaleacc}. We can see that it works well over a wide range of values. Note that $\alpha=1$ case has been discussed by SPC1 and mSPC1 methods in Table 2. When $\alpha=1$, Eq. (\ref{obj}) and (\ref{multimodel}) do not possess similarity preserving capability. 

\section{Conclusion}
In this paper, we propose a clustering algorithm which can exploit similarity information of raw data. Furthermore, the structure information of a graph is also considered in our objective function. Comprehensive experimental results on real data sets well demonstrate the superiority of the proposed method on the clustering task. It has been shown that the performance of the proposed method is largely determined by the choice of the kernel function. To this end, we develop a multiple kernel learning method, which is capable of automatically learning an appropriate kernel from a pool of candidate kernels. In the future, we will examine the effectiveness of our framework on the semi-supervised learning task.
\section*{Acknowledgment}
This paper was in part supported by Grants from the Natural
Science Foundation of China (Nos. 61806045, 61572111, and 61872062), three Fundamental Research Fund for the Central Universities of China (Nos. ZYGX2017KYQD177, A03017023701012, and ZYGX2016J086), and a 985 Project of UESTC (No. A1098531023601041).
\section*{References}
\bibliographystyle{elsarticle-num} 
\bibliography{cvprref}


\end{document}